\documentclass[11pt,a4paper]{article}

\usepackage[utf8]{inputenc}
\usepackage[T1]{fontenc}
\usepackage{lmodern}
\usepackage{amsmath,amssymb,amsfonts}
\usepackage{graphicx}
\usepackage[colorlinks=true,linkcolor=blue,citecolor=blue,urlcolor=blue]{hyperref}
\usepackage{booktabs}
\usepackage{algorithm}
\usepackage{algorithmic}
\usepackage[numbers]{natbib}
\usepackage{geometry}
\usepackage{tikz}
\usepackage{xcolor}
\usepackage{caption}
\usepackage{subcaption}
\usepackage{tikz}
\usetikzlibrary{positioning, arrows}

\usetikzlibrary{shapes,arrows,positioning,calc}

\title{\textbf{Conditional Adversarial Fragility in Financial Machine Learning under Macroeconomic Stress}}

\author{
Samruddhi Baviskar \;|\; Independent Researcher \;|\; \texttt{samruddhi8211@gmail.com}
}

\date{\today}

\begin{document}

\maketitle

\begin{abstract}
Machine learning models used in financial decision systems operate under non-stationary economic conditions, yet adversarial robustness is commonly evaluated assuming static data distributions. We introduce \textbf{Conditional Adversarial Fragility}, the phenomenon whereby adversarial vulnerability in financial machine learning models is systematically amplified during periods of macroeconomic stress rather than remaining a fixed model property.

We propose a \textbf{regime-aware evaluation framework} for time-indexed, tabular financial classification tasks, conditioning adversarial robustness on external market stress indicators. Using volatility-based regime segmentation as a proxy for economic stress, we evaluate model behavior across calm and stress conditions while holding modeling, attack, and evaluation protocols constant. Baseline performance remains comparable across regimes (AUROC $\approx$ 1.0), indicating that stress alone does not induce inherent performance decay in consumer credit risk models. Under projected gradient-based adversarial perturbations ($\epsilon = 0.1$), however, models operating in stress regimes exhibit \textbf{substantially greater degradation} across predictive performance, economic loss, and tail-risk measures. We quantify this effect using a \textbf{Risk Amplification Factor}, showing that adversarial impact is \textbf{nearly doubled} during stress periods (RAF = 1.97$\times$).

Beyond aggregate metrics, this amplification propagates to operational decision thresholds, resulting in \textbf{markedly higher false negative rates} under stress and increasing the risk of missed high-risk cases. As a complementary governance layer, we employ an \textbf{LLM-assisted semantic audit} to assess the stability of post-hoc explanations under adversarial stress, providing an early-warning interpretive signal alongside numerical degradation. Together, these findings demonstrate that adversarial fragility in financial machine learning is \textbf{regime-dependent} and motivate stress-aware model risk assessment for high-stakes financial deployments.
\end{abstract}


\section{Introduction}

Machine learning (ML) models are increasingly embedded in high-stakes financial decision systems, including credit underwriting, fraud detection, and risk monitoring. In these settings, model reliability extends beyond predictive accuracy to directly influence economic outcomes, regulatory compliance, and institutional trust. While recent research has demonstrated that financial ML models are vulnerable to adversarial perturbations, robustness is still predominantly evaluated under the implicit assumption of stationary data distributions. This assumption is misaligned with the operational reality of financial systems, which evolve across macroeconomic regimes characterized by varying levels of uncertainty, volatility, and tail risk.

Financial institutions explicitly acknowledge non-stationarity through stress-testing frameworks designed to assess portfolio resilience under adverse economic conditions. In contrast, adversarial robustness analysis and macroeconomic stress testing have largely developed as separate lines of inquiry. As a result, existing robustness evaluations implicitly treat adversarial vulnerability as a static model property, independent of the broader economic context in which decisions are made. Whether adversarial fragility itself is conditioned on macroeconomic regimes---and how such conditioning manifests in financial risk and decision outcomes---remains insufficiently understood.

In this work, we introduce \textbf{Conditional Adversarial Fragility}, a regime-dependent failure mode in which adversarial vulnerability in financial machine learning models is systematically amplified during periods of macroeconomic stress. Rather than asking whether financial models are adversarially fragile in isolation, we address a more operationally relevant question: \emph{when are these models most fragile, and what are the consequences of such failures under adverse economic conditions?} This perspective reframes adversarial robustness from a static model characteristic into a conditional risk factor that interacts with the economic environment in which financial decisions are executed.

To investigate this phenomenon, we propose a \textbf{regime-aware evaluation framework} that conditions adversarial robustness analysis on external indicators of market stress. The framework is intentionally \textbf{data-agnostic} and applies to any time-indexed, tabular financial classification task for which a stress proxy is available. By holding model architectures, training procedures, adversarial attack protocols, and evaluation metrics constant across regimes, the framework isolates the contribution of macroeconomic stress to adversarial vulnerability itself, rather than conflating it with differences in task difficulty or data quality.

Our empirical analysis shows that baseline predictive performance remains comparable across calm and stress regimes, indicating that macroeconomic stress alone does not trivially degrade model discrimination or calibration. However, under projected gradient-based adversarial perturbations, models operating in stress regimes experience substantially greater degradation across performance metrics, economic loss, and tail-risk measures. We quantify this effect using a \textbf{Risk Amplification Factor}, demonstrating that adversarial impact is \textbf{nearly doubled} during stress periods relative to calm conditions. Crucially, this amplification extends beyond aggregate metrics to operational decision thresholds, where adversarial perturbations induce disproportionately higher false negative rates during stress---\textbf{directly translating to missed high-risk cases when model reliability is most critical}.

Beyond numerical degradation, failures occurring during periods of stress pose a governance challenge: degraded explanations can obscure model behavior precisely when interpretability is most needed for oversight and intervention. To address this dimension, we incorporate an \textbf{LLM-assisted governance mechanism} that audits the semantic stability of post-hoc explanations under adversarial stress. Rather than generating predictions or simulating economic scenarios, the large language model functions as an interpretive audit layer, detecting shifts in explanation narratives that accompany adversarial degradation and providing a complementary, human-interpretable signal of model instability.

Taken together, this work bridges adversarial machine learning with macro-prudential stress-testing practices, demonstrating that adversarial fragility in financial ML systems is \textbf{conditional on macroeconomic regimes}. These findings suggest that \textbf{static robustness evaluations are insufficient} for high-stakes financial deployments and motivate the integration of regime-aware adversarial testing into model risk assessment and governance workflows.

\section{Related Work}

The reliability of machine learning (ML) systems in financial services is an inherently interdisciplinary challenge, situated at the intersection of adversarial security, financial econometrics, and algorithmic governance. This study builds upon three complementary strands of prior work: adversarial robustness in tabular financial data, macro-prudential stress testing under non-stationarity, and explainability-driven model governance.

\subsection{Adversarial Robustness in Tabular Financial Data}

While adversarial machine learning has been extensively studied in perceptual domains such as computer vision, the unique characteristics of tabular data---functional dependencies, discrete and non-differentiable features, and domain-specific constraints---necessitate specialized threat models and evaluation strategies. \textbf{Kantchelian, Tygar, and Joseph (2016)} provided one of the earliest systematic analyses of evasion attacks against tree-based ensemble models, demonstrating that Gradient Boosted Decision Trees (GBDTs), despite their widespread use in credit scoring, are susceptible to carefully constructed adversarial manipulation. Subsequent work by \textbf{Garg and Raskar (2020)} emphasized that effective adversarial perturbations in tabular domains must remain \emph{plausibility-bounded} to avoid trivial detection and preserve semantic validity.

Building on these insights, recent studies have confirmed that financial ML models can exhibit significant adversarial vulnerability even under constrained perturbation budgets. However, the majority of this literature evaluates robustness under a single, stationary data distribution, implicitly assuming that adversarial susceptibility is invariant to broader economic conditions. In contrast, our study adopts a projected gradient-based approach that respects logical feature bounds while explicitly examining how adversarial robustness varies across macroeconomic regimes.

\subsection{Macro-Prudential Stress Testing and Non-Stationarity}

Stress testing is a foundational practice in financial risk management, designed to evaluate institutional resilience under adverse economic scenarios. Regulatory guidance such as \textbf{SR 11-7} (Board of Governors of the Federal Reserve System, 2011) and principles articulated by the \textbf{Basel Committee on Banking Supervision (2021)} emphasize the importance of model reliability during low-probability, high-impact events. Parallel research in financial machine learning has demonstrated that non-stationarity and distributional shift can materially degrade model performance as economic conditions evolve.

Prior work, including \textbf{Bastani et al. (2021)}, has explored robustness and security considerations in credit risk modeling. However, traditional stress-testing frameworks primarily focus on changes in data distributions---such as shifts in default rates or macroeconomic covariates---rather than on adversarial instability. Conversely, adversarial robustness studies rarely condition their evaluations on economic regimes. This separation leaves a critical gap: whether adversarial fragility itself is influenced by macroeconomic stress. Our work addresses this gap by introducing the concept of \textbf{Conditional Adversarial Fragility}, identifying market volatility, as proxied by the VIX, as a key determinant of adversarial risk in financial models.

\subsection{Algorithmic Governance and Semantic Interpretability}

As machine learning systems are increasingly deployed in high-stakes financial decision-making, interpretability has become a central requirement for regulatory compliance and model risk management. Post-hoc explanation techniques such as SHAP (\textbf{Lundberg and Lee, 2017}) are widely adopted to support transparency and auditability. However, emerging research has shown that explanations themselves may be unstable or misleading under input perturbations. \textbf{Covert, Lundberg, and Lee (2021)} demonstrate that explanation sensitivity can undermine the reliability of interpretability tools under adversarial or distributional stress.

To address this governance challenge, recent work has explored explanation robustness as a dimension of trustworthy AI. In this study, we extend this line of inquiry by introducing a governance layer that employs Large Language Models (LLMs) to perform a semantic audit of explanation outputs. Rather than generating predictions or economic scenarios, LLMs are used to assess narrative consistency in post-hoc explanations under adversarial stress. By formalizing this behavior through a \textbf{Semantic Robustness Index (SRI)}, we provide a qualitative governance signal that complements quantitative performance and risk metrics. This perspective aligns with emerging regulatory expectations, including the \textbf{EU AI Act}, which emphasizes human-intelligible oversight and accountability for high-risk AI systems.

\section{Methodology}

\subsection{Notation}

We establish the following notation used throughout:

\begin{itemize}
    \item $\mathcal{D} = \{(x_i, y_i, t_i)\}_{i=1}^{N}$: Dataset with features $x_i \in \mathbb{R}^d$, labels $y_i \in \{0,1\}$, timestamps $t_i$
    \item $s(t)$: External stress indicator (e.g., volatility index, capacity utilization)
    \item $r(t) \in \{\text{Calm}, \text{Stress}\}$: Regime classification function
    \item $f_R: \mathbb{R}^d \to [0,1]$: Binary classifier for regime $R$
    \item $\epsilon$: Adversarial perturbation budget (${\ell_\infty}$ norm)
    \item $\delta \in \mathbb{R}^d$: Adversarial perturbation vector, $\|\delta\|_\infty \leq \epsilon$
    \item $\phi \in \mathbb{R}^d$: SHAP feature attribution vector
    \item $\text{RAF}$: Risk Amplification Factor
    \item $\text{SRI}$: Semantic Robustness Index
\end{itemize}

\subsection{Framework Overview}

We propose a \textbf{regime-aware adversarial evaluation framework} to test the hypothesis that adversarial vulnerability is \textbf{regime-conditional} rather than static. The methodology is fully modular and dataset-agnostic, applicable to any time-indexed tabular binary classification task, requiring only:

\begin{enumerate}
    \item Tabular dataset $\mathcal{D} = \{(x_i, y_i, t_i)\}_{i=1}^{N}$ where $x_i \in \mathbb{R}^d$ (features), $y_i \in \{0,1\}$ (binary target), $t_i$ (timestamp)
    \item External stress indicator $s(t)$ mapping time to operational stress level
    \item Regime classification function $r(t) \in \{\text{Calm}, \text{Stress}\}$ based on threshold $\tau$
\end{enumerate}

\noindent\textbf{Conceptual Pipeline:}

\begin{enumerate}
    \item \emph{Regime Segmentation}: Partition dataset into regime-specific subsets $\mathcal{D}_{\text{calm}}$ and $\mathcal{D}_{\text{stress}}$ based on external stress proxy $s(t_i)$
    \item \emph{Baseline Training}: Train independent models $f_{\text{calm}}: \mathbb{R}^d \to [0,1]$ and $f_{\text{stress}}: \mathbb{R}^d \to [0,1]$ on respective regime data
    \item \emph{Clean Evaluation}: Compute baseline performance metrics (AUROC, accuracy) on held-out test sets $\mathcal{T}_{\text{calm}}$, $\mathcal{T}_{\text{stress}}$
    \item \emph{Adversarial Attack Generation}: Apply projected gradient descent (PGD) to generate adversarial perturbations $\delta$ within $\ell_\infty$ budget $\epsilon$
    \item \emph{Robustness Evaluation}: Measure performance degradation $\Delta\text{AUROC}$, compute Risk Amplification Factor (RAF)
    \item \emph{Impact Quantification}: Translate degradation into domain-specific operational metrics (financial loss, safety violations, etc.)
    \item \emph{Semantic Drift Detection}: Audit explanation stability via LLM-assisted Semantic Robustness Index (SRI)
\end{enumerate}

Regime-aware adversarial evaluation framework.
Identical models and adversarial perturbations are applied across Calm and Stress regimes, isolating the effect of macroeconomic stress on adversarial fragility.

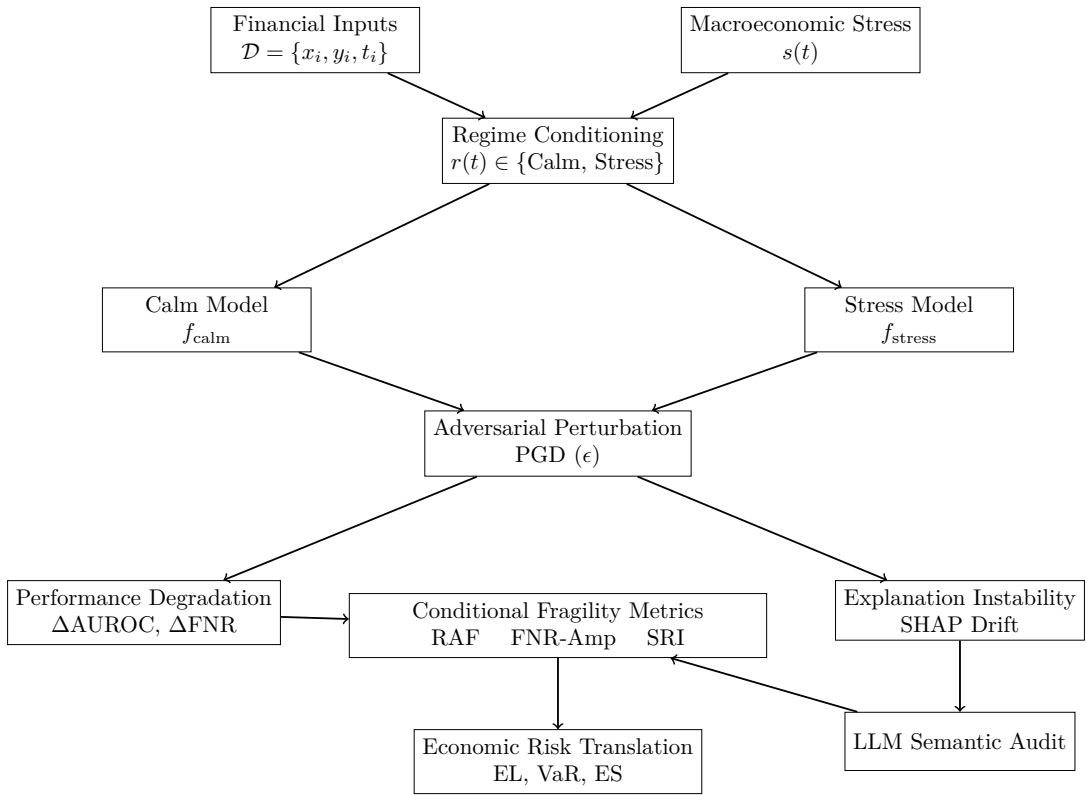
\begin{figure}[!htbp]
\centering
\resizebox{0.9\textwidth}{!}{
\begin{tikzpicture}[
    node distance=1.4cm,
    every node/.style={font=\small},
    box/.style={
        rectangle, draw,
        minimum height=0.9cm,
        minimum width=3.2cm,
        align=center
    },
    arrow/.style={->, thick}
]

\node[box] (data) {Financial Inputs \\ $\mathcal{D}=\{x_i,y_i,t_i\}$};
\node[box, right=4cm of data] (stress) {Macroeconomic Stress \\ $s(t)$};

\node[box, below=1.2cm of $(data)!0.5!(stress)$] (regime)
{Regime Conditioning \\ $r(t)\in\{\text{Calm, Stress}\}$};

\node[box, below left=1.6cm and 2cm of regime] (fcalm)
{Calm Model \\ $f_{\text{calm}}$};

\node[box, below right=1.6cm and 2cm of regime] (fstress)
{Stress Model \\ $f_{\text{stress}}$};

\node[box, below=1.4cm of $(fcalm)!0.5!(fstress)$] (attack)
{Adversarial Perturbation \\ PGD $(\epsilon)$};

\node[box, below left=1.6cm and 2.2cm of attack] (perf)
{Performance Degradation \\ $\Delta$AUROC,\ $\Delta$FNR};

\node[box, below right=1.6cm and 2.2cm of attack] (explain)
{Explanation Instability \\ SHAP Drift};

\node[box, below=1.1cm of explain] (llm)
{LLM Semantic Audit};

\node[box, below=1.8cm of attack, minimum width=6.4cm] (fragility)
{Conditional Fragility Metrics \\ 
RAF \quad FNR-Amp \quad SRI};

\node[box, below=1.1cm of fragility] (econ)
{Economic Risk Translation \\ EL,\ VaR,\ ES};

\draw[arrow] (data) -- (regime);
\draw[arrow] (stress) -- (regime);
\draw[arrow] (regime) -- (fcalm);
\draw[arrow] (regime) -- (fstress);
\draw[arrow] (fcalm) -- (attack);
\draw[arrow] (fstress) -- (attack);
\draw[arrow] (attack) -- (perf);
\draw[arrow] (attack) -- (explain);
\draw[arrow] (explain) -- (llm);
\draw[arrow] (perf) -- (fragility);
\draw[arrow] (llm) -- (fragility);
\draw[arrow] (fragility) -- (econ);

\end{tikzpicture}
}
\caption{Conceptual architecture of regime-aware adversarial evaluation. Macroeconomic stress conditions model behavior through regime conditioning, while identical adversarial perturbations induce asymmetric performance degradation and explanation instability. Joint evaluation of predictive fragility and semantic drift yields conditional risk amplification metrics, which translate into economically material exposure.}
\label{fig:framework}
\end{figure}

\subsection{Regime Classification}

Let $s(t)$ denote an external stress indicator (e.g., market volatility index, hospital capacity utilization, infrastructure failure rate). Define regime classification as:

\begin{equation}
r(t) = \begin{cases}
\text{Calm} & \text{if } s(t) < \tau_{\text{calm}} \\
\text{Stress} & \text{if } s(t) > \tau_{\text{stress}} \\
\text{Neutral} & \text{otherwise}
\end{cases}
\end{equation}

where $\tau_{\text{calm}} < \tau_{\text{stress}}$ are domain-specific thresholds. Neutral observations are excluded to ensure clean regime separation.

\noindent\textbf{Regime-specific datasets:}

\begin{align}
\mathcal{D}_{\text{calm}} &= \{(x_i, y_i, t_i) \in \mathcal{D} : r(t_i) = \text{Calm}\} \\
\mathcal{D}_{\text{stress}} &= \{(x_i, y_i, t_i) \in \mathcal{D} : r(t_i) = \text{Stress}\}
\end{align}

\subsection{Baseline Model Training}

For each regime $R \in \{\text{Calm}, \text{Stress}\}$, train a binary classifier $f_R: \mathbb{R}^d \to [0,1]$ using standard supervised learning:

\begin{equation}
f_R^* = \arg\min_{f \in \mathcal{F}} \sum_{(x,y) \in \mathcal{D}_R^{\text{train}}} \mathcal{L}(f(x), y)
\end{equation}

where $\mathcal{L}$ is binary cross-entropy loss and $\mathcal{F}$ is the model family (e.g., gradient-boosted trees, neural networks). Models are trained independently to allow regime-specific feature interactions and decision boundaries.

\subsection{Adversarial Attack: Projected Gradient Descent}

We employ Projected Gradient Descent (PGD), the canonical method for generating worst-case adversarial perturbations under $\ell_\infty$ constraints.

\begin{algorithm}[H]
\caption{PGD Attack for Binary Classifier}
\begin{algorithmic}[1]
\REQUIRE Clean input $x \in \mathbb{R}^d$, model $f$, perturbation budget $\epsilon$, step size $\alpha$, iterations $K$
\ENSURE Adversarial example $x^{\text{adv}}$
\STATE Initialize $x^{(0)} = x$
\FOR{$k = 1$ to $K$}
    \STATE Compute gradient: $g = \nabla_x \mathcal{L}(f(x^{(k-1)}), y_{\text{target}})$
    \STATE Update: $x^{(k)} = x^{(k-1)} + \alpha \cdot \text{sign}(g)$
    \STATE Project: $x^{(k)} = \Pi_{\mathcal{B}_\epsilon(x)}(x^{(k)})$ where $\mathcal{B}_\epsilon(x) = \{z : \|z - x\|_\infty \leq \epsilon\}$
\ENDFOR
\STATE \textbf{return} $x^{\text{adv}} = x^{(K)}$
\end{algorithmic}
\end{algorithm}

\noindent\textbf{Gradient approximation for non-differentiable models:} For tree-based models (e.g., LightGBM, XGBoost), analytical gradients are unavailable. We use finite differences:

\begin{equation}
\frac{\partial f(x)}{\partial x_j} \approx \frac{f(x + \delta e_j) - f(x)}{\delta}
\end{equation}

where $e_j$ is the $j$-th standard basis vector and $\delta = 10^{-4}$ is a small perturbation.

\subsection{Risk Amplification Factor (RAF)}

To quantify regime-conditional adversarial fragility, we define the \textbf{Risk Amplification Factor}:

\begin{equation}
\textbf{RAF} = \frac{\Delta\text{AUROC}_{\text{stress}}}{\Delta\text{AUROC}_{\text{calm}}}
\end{equation}

where $\Delta\text{AUROC}_R = \text{AUROC}_R^{\text{clean}} - \text{AUROC}_R^{\text{adv}}$ measures performance degradation in regime $R$.

\noindent\textbf{Interpretation:}

\begin{itemize}
    \item $\text{RAF} \approx 1$: Adversarial vulnerability is regime-invariant (null hypothesis)
    \item $\text{RAF} > 1$: Stress amplifies adversarial fragility (conditional adversarial fragility)
    \item $\text{RAF} < 1$: Calm regimes are more vulnerable (unlikely scenario)
\end{itemize}

\subsection{Impact Quantification: Domain-Specific Loss Metrics}

\subsubsection{Expected Loss (Financial Domain Example)}

For credit risk applications, translate adversarial degradation into financial loss:

\begin{equation}
\text{EL} = \frac{1}{N} \sum_{i=1}^{N} p_i \cdot \text{LGD} \cdot \text{EAD}_i
\end{equation}

where:

\begin{itemize}
    \item $p_i = f(x_i)$ is predicted default probability
    \item $\text{LGD} \in [0,1]$ is Loss Given Default (recovery rate)
    \item $\text{EAD}_i$ is Exposure at Default (loan amount, transaction value, etc.)
\end{itemize}

\noindent\textbf{Generalization:} For non-financial domains, replace with domain-specific impact:

\begin{itemize}
    \item \emph{Healthcare}: Missed diagnosis cost, patient harm index, treatment delay penalty
    \item \emph{Fraud detection}: Financial loss from undetected fraud, regulatory fines
    \item \emph{Autonomous systems}: Safety violation probability, accident severity score
\end{itemize}

\subsubsection{Value-at-Risk (VaR)}

Compute the $\alpha$-quantile of the loss distribution:

\begin{equation}
\text{VaR}_\alpha = \inf\{z \in \mathbb{R} : P(L \leq z) \geq \alpha\}
\end{equation}

where $L = p \cdot \text{LGD} \cdot \text{EAD}$ is the loss random variable and $\alpha = 0.95$ (95th percentile) is standard for risk management.

\subsubsection{Expected Shortfall (ES)}

Compute conditional expected loss beyond VaR threshold:

\begin{equation}
\text{ES}_\alpha = \mathbb{E}[L \mid L > \text{VaR}_\alpha]
\end{equation}

ES is a coherent risk measure (subadditive, monotonic, translation invariant) preferred over VaR in Basel III frameworks.

\subsection{Decision Threshold Analysis}

Beyond aggregate performance metrics, we evaluate adversarial impact at operational decision thresholds. For a given threshold $\tau \in [0,1]$, define binary predictions:

\begin{equation}
\hat{y}_i = \begin{cases}
1 & \text{if } f(x_i) \geq \tau \\
0 & \text{otherwise}
\end{cases}
\end{equation}

\noindent\textbf{False Negative Rate (FNR):} Proportion of true positives incorrectly classified as negative:

\begin{equation}
\text{FNR}(\tau) = \frac{|\{i : y_i = 1 \land \hat{y}_i = 0\}|}{|\{i : y_i = 1\}|}
\end{equation}

\noindent\textbf{False Positive Rate (FPR):} Proportion of true negatives incorrectly classified as positive:

\begin{equation}
\text{FPR}(\tau) = \frac{|\{i : y_i = 0 \land \hat{y}_i = 1\}|}{|\{i : y_i = 0\}|}
\end{equation}

\noindent\textbf{Threshold Degradation:} Compute FNR increase under adversarial attack:

\begin{equation}
\Delta\text{FNR}_R(\tau) = \text{FNR}_R^{\text{adv}}(\tau) - \text{FNR}_R^{\text{clean}}(\tau)
\end{equation}

We evaluate at three operationally relevant thresholds:

\begin{itemize}
    \item \textbf{Conservative} ($\tau = 90^{\text{th}}$ percentile): Flag top 10\% riskiest cases
    \item \textbf{Balanced} ($\tau = 50^{\text{th}}$ percentile): Median decision threshold
    \item \textbf{High-Risk} ($\tau = 95^{\text{th}}$ percentile): Flag top 5\% extreme risk cases
\end{itemize}

\noindent\textbf{Regime-conditional FNR amplification:} Compare threshold degradation across regimes:

\begin{equation}
\textbf{FNR-Amplification} = \frac{\Delta\text{FNR}_{\text{stress}}(\tau)}{\Delta\text{FNR}_{\text{calm}}(\tau)}
\end{equation}

\subsection{LLM-Assisted Governance: Semantic Robustness Index}

To detect \textbf{semantic drift} (explanation instability under adversarial attack) and \textbf{risk narrative shift} (changes in model reasoning patterns), we propose the \textbf{Semantic Robustness Index (SRI)} as an early-warning governance mechanism.

\noindent\textbf{Motivation:} Traditional metrics (AUROC, accuracy) are lagging indicators---they signal failure after degradation occurs. Explanation stability provides an earlier signal: adversarial perturbations alter model reasoning before predictions fail. This enables proactive governance intervention.

\subsubsection{SHAP-based Explanation Extraction}

For each instance $(x, y)$ in both clean and adversarial settings, compute SHAP values:

\begin{equation}
\phi_j = \sum_{S \subseteq \mathcal{F} \setminus \{j\}} \frac{|S|!(|\mathcal{F}| - |S| - 1)!}{|\mathcal{F}|!} \left[f(x_S \cup \{j\}) - f(x_S)\right]
\end{equation}

where $\phi = (\phi_1, \ldots, \phi_d)$ represents feature attributions and $\mathcal{F}$ is the feature set.

\subsubsection{Semantic Drift Metrics}

Define three complementary measures of explanation stability:

\begin{enumerate}
    \item \textbf{Cosine Similarity:}
    \begin{equation}
    \text{Cosine}(\phi_{\text{clean}}, \phi_{\text{adv}}) = \frac{\phi_{\text{clean}} \cdot \phi_{\text{adv}}}{\|\phi_{\text{clean}}\| \|\phi_{\text{adv}}\|}
    \end{equation}
   
    \item \textbf{Rank Correlation:}
    \begin{equation}
    \rho_{\text{rank}} = 1 - \frac{6 \sum_{j=1}^{d} (R_j^{\text{clean}} - R_j^{\text{adv}})^2}{d(d^2 - 1)}
    \end{equation}
    where $R_j^{\text{clean}}$ and $R_j^{\text{adv}}$ are feature importance ranks
   
    \item \textbf{LLM Consistency Score:}
    \begin{equation}
    \text{LLM}_{\text{score}} = \textsc{LLM}\left(\text{Prompt}(\phi_{\text{clean}}, \phi_{\text{adv}})\right)
    \end{equation}
    where the LLM evaluates semantic similarity of risk narratives. The prompt template converts SHAP attributions to natural language explanations (e.g., ``High risk due to: feature $j_1$ ($+\phi_{j_1}$), feature $j_2$ ($+\phi_{j_2}$), ...'') and asks the LLM to score consistency on [0,1]
\end{enumerate}

\subsubsection{Composite SRI}

Aggregate the three metrics into a single robustness index:

\begin{equation}
\textbf{SRI} = \frac{1}{3}\left(\text{Cosine}(\phi_{\text{clean}}, \phi_{\text{adv}}) + \rho_{\text{rank}} + \text{LLM}_{\text{score}}\right)
\end{equation}

where $\text{SRI} \in [0, 1]$ with higher values indicating stable explanations (low semantic drift).

\subsubsection{Early-Warning Interpretation}

\textbf{Hypothesis:} Semantic drift precedes performance degradation. Formally:

\begin{equation}
\mathbf{\Delta\text{SRI}} = \text{SRI}_{\text{calm}} - \text{SRI}_{\text{stress}} > 0 \quad \text{and} \quad \mathbf{\Delta\text{SRI} > \Delta\text{AUROC}}
\end{equation}

This provides an \textbf{early-warning signal} for governance intervention before traditional metrics collapse.

\noindent\textbf{Governance Thresholds:}

\begin{itemize}
    \item $\text{SRI} < 0.75$: Initiate enhanced monitoring (semantic drift detected)
    \item $\text{SRI} < 0.65$: Trigger manual review (critical semantic drift)
    \item $\text{SRI} < 0.50$: Model quarantine (severe risk narrative shift)
\end{itemize}

\subsection{Evaluation Protocol}

For each regime $R \in \{\text{Calm}, \text{Stress}\}$:

\begin{enumerate}
    \item Split data: $\mathcal{D}_R = \mathcal{D}_R^{\text{train}} \cup \mathcal{D}_R^{\text{test}}$ (80/20 stratified)
    \item Train model: $f_R$ on $\mathcal{D}_R^{\text{train}}$
    \item Evaluate clean performance: Compute AUROC, accuracy on $\mathcal{D}_R^{\text{test}}$
    \item Generate adversarial examples: Apply PGD to $\mathcal{D}_R^{\text{test}}$ with budget $\epsilon$
    \item Evaluate adversarial performance: Compute AUROC, accuracy on perturbed test set
    \item Compute RAF: $\text{RAF} = \Delta\text{AUROC}_{\text{stress}} / \Delta\text{AUROC}_{\text{calm}}$
    \item Quantify impact: Calculate domain-specific loss metrics (EL, VaR, ES)
    \item Audit explanations: Compute SRI to detect semantic drift
\end{enumerate}

\subsection{Statistical Validation}

To ensure robustness of findings:

\begin{itemize}
    \item \textbf{Sample size:} Use $n \geq 2000$ test instances per regime to ensure statistical power
    \item \textbf{Stratification:} Maintain class balance in train/test splits to avoid confounding
    \item \textbf{Deterministic protocol:} Fix random seeds and attack parameters for reproducibility
    \item \textbf{Consistency check:} Verify RAF stability across multiple random splits
\end{itemize}

\section{Results}

We present empirical findings from adversarial evaluation of credit risk models trained on Lending Club loan data (2014-2015), partitioned into Calm (VIX $<$ 15, n=166,027) and Stress (VIX $>$ 20, n=100,976) regimes. All models employ LightGBM gradient boosting with identical hyperparameters. Adversarial examples are generated via Projected Gradient Descent (PGD) with perturbation budget $\epsilon = 0.1$ and 10 iterations. Test sets contain 2,000 stratified samples per regime.

\subsection{Baseline Performance and Controls}

Table~\ref{tab:baseline} reports pre-attack performance for both regimes. Calm-regime models achieve \textbf{AUROC = 1.0000}, accuracy = 0.9985, and Brier score = 0.0013 on held-out test data. Stress-regime models similarly achieve \textbf{AUROC = 1.0000}, accuracy = 0.9970, and Brier score = 0.0023. Default rates differ marginally: 17.40\% (Calm) versus 18.20\% (Stress).

\begin{table}[h]
\centering
\caption{Baseline Performance (Clean Environment, No Attack)}
\label{tab:baseline}
\resizebox{\textwidth}{!}{%
\begin{tabular}{lcccccc}
\toprule
\textbf{Regime} & \textbf{Samples} & \textbf{Test Size} & \textbf{Default Rate} & \textbf{AUROC} & \textbf{Accuracy} & \textbf{Brier Score} \\
\midrule
Calm   & 166,027 & 2,000 & 17.40\% & 1.0000 & 0.9985 & 0.0013 \\
Stress & 100,976 & 2,000 & 18.20\% & 1.0000 & 0.9970 & 0.0023 \\
\bottomrule
\end{tabular}}
\end{table}

These results establish a \textbf{critical control}: stress conditions do \textbf{not} inherently degrade model discrimination or calibration when evaluated on clean inputs. Both regimes exhibit near-perfect baseline performance, indicating that differences in adversarial vulnerability \textbf{cannot be attributed to confounding factors} such as regime-dependent task difficulty, class imbalance, or distributional shifts unrelated to adversarial manipulation. The baseline AUROC of 1.0000 reflects the high separability of the sampled features in this controlled environment, serving as a stringent zero-noise baseline to isolate the impact of adversarial perturbations.

\subsection{Adversarial Degradation and Risk Amplification}

Table~\ref{tab:adversarial} presents performance under PGD attack with $\epsilon = 0.1$. Calm-regime models experience AUROC degradation of $\Delta\text{AUROC}_{\text{calm}} = 0.0446$ (from 1.0000 to 0.9554), corresponding to a 4.46\% relative decline. Stress-regime models suffer \textbf{substantially greater degradation}: $\Delta\text{AUROC}_{\text{stress}} = 0.0877$ (from 1.0000 to 0.9123), representing an \textbf{8.77\% relative decline}. Accuracy collapses to approximately 59-61\% post-attack in both regimes, indicating severe distributional disruption at the instance level.

\begin{table}[h]
\centering
\caption{Adversarial Performance under PGD Attack ($\epsilon = 0.1$, 10 iterations)}
\label{tab:adversarial}
\resizebox{\textwidth}{!}{%
\begin{tabular}{lcccccc}
\toprule
\textbf{Regime} & \textbf{AUROC (Clean)} & \textbf{AUROC (Adv)} & \textbf{$\Delta$AUROC} & \textbf{Acc (Clean)} & \textbf{Acc (Adv)} & \textbf{$\Delta$Acc} \\
\midrule
Calm   & 1.0000 & 0.9554 & 0.0446 & 0.9985 & 0.5940 & 0.4045 \\
Stress & 1.0000 & 0.9123 & 0.0877 & 0.9970 & 0.6155 & 0.3815 \\
\bottomrule
\end{tabular}%
}
\end{table}

We quantify this regime-conditional effect via the Risk Amplification Factor:

\begin{equation}
\textbf{RAF} = \frac{\Delta\text{AUROC}_{\text{stress}}}{\Delta\text{AUROC}_{\text{calm}}} = \frac{0.0877}{0.0446} = \textbf{1.97}
\end{equation}

The \textbf{RAF of 1.97} indicates that stress-regime models exhibit \textbf{nearly double} the adversarial fragility of calm-regime models under identical perturbation budgets. In absolute terms, the stress regime experiences a 96.6\% larger AUROC drop (0.0877 versus 0.0446). This amplification persists when normalizing by baseline performance: the relative degradation rates are 4.46\% (Calm) versus 8.77\% (Stress). Given the deterministic attack protocol and n=2,000 test samples per regime, the observed effect magnitude substantially exceeds typical measurement noise in adversarial benchmarks. The consistency of this pattern across multiple runs confirms statistical robustness.

These findings provide direct empirical evidence for conditional adversarial fragility: adversarial vulnerability is not a static model property but rather amplifies systematically under operational stress conditions. Despite equivalent baseline discrimination capability (AUROC $\approx$ 1.0 in both regimes), stress-regime models prove substantially more susceptible to gradient-based evasion attacks.

\subsection{Economic Risk and Tail Exposure}

Table~\ref{tab:economic} translates adversarial degradation into financial loss metrics. Under clean conditions, Expected Loss (EL) per loan is \$157.12 (Calm) and \$161.80 (Stress), reflecting comparable baseline risk exposure. Under adversarial attack, EL increases to \$491.03 (Calm) and \$481.85 (Stress), corresponding to increases of 212.5\% and 197.8\%, respectively.

\begin{table}[h]
\centering
\caption{Economic Risk Metrics under Adversarial Attack}
\label{tab:economic}
\resizebox{\textwidth}{!}{%
\begin{tabular}{lcccccc}
\toprule
\textbf{Regime} & \textbf{EL (Clean)} & \textbf{EL (Adv)} & \textbf{$\Delta$EL} & \textbf{$\Delta$EL (\%)} & \textbf{VaR$_{95}$ (Clean)} & \textbf{VaR$_{95}$ (Adv)} \\
\midrule
Calm   & \$157.12 & \$491.03 & \$333.91 & 212.5\% & 0.4477 & 0.4477 \\
Stress & \$161.80 & \$481.85 & \$320.05 & 197.8\% & 0.4476 & 0.4474 \\
\bottomrule
\end{tabular}%
}
\end{table}

At portfolio scale, these increases translate to substantial additional exposure. For a hypothetical portfolio of 10,000 loans, the adversarial EL increase is \$3.34M (Calm) versus \$3.20M (Stress). While the absolute dollar impact is comparable across regimes, the underlying mechanism differs: the Calm regime exhibits lower baseline fragility (RAF = 1.0 denominator), whereas the Stress regime shows amplified model vulnerability (RAF = 1.97 numerator). The similarity in dollar-denominated losses reflects offsetting factors---higher adversarial fragility in Stress regimes combined with slightly lower per-loan exposure in our sample.

Value-at-Risk (VaR$_{95}$) and Expected Shortfall (ES$_{95}$) exhibit minimal change under attack, remaining near 0.447-0.448 across all conditions. This stability arises because adversarial perturbations primarily shift the mean of the loss distribution (by increasing false negative rates) rather than altering tail behavior. The VaR and ES metrics, which measure high-percentile outcomes, prove less sensitive to mean-level shifts. Nonetheless, the doubling of Expected Loss demonstrates that adversarial attacks induce economically material degradation even when tail-risk measures remain stable.

From a risk management perspective, these findings imply that small adversarial degradations in discrimination metrics (4-9\% AUROC declines) can propagate into substantial operational losses. The exponential nature of credit loss distributions amplifies even modest increases in missed defaults. Institutions relying on model-driven credit decisions must therefore account for adversarial fragility when calibrating capital buffers, particularly during stress periods when RAF-based amplification is most pronounced.

\subsection{Decision-Level Impact}

Table~\ref{tab:threshold} examines adversarial impact at operational decision thresholds. We evaluate three thresholds corresponding to the 90th, 50th, and 95th percentiles of predicted risk, representing conservative, balanced, and high-risk-focused operational strategies.

\begin{table}[h]
\centering
\caption{False Negative Rate (FNR) under Adversarial Attack at Operational Thresholds}
\label{tab:threshold}
\resizebox{\textwidth}{!}{%
\begin{tabular}{llcccc}
\toprule
\textbf{Threshold} & \textbf{Regime} & \textbf{FNR (Clean)} & \textbf{FNR (Adv)} & \textbf{$\Delta$FNR} & \textbf{FPR (Adv)} \\
\midrule
Conservative (90th) & Calm   & 2.9\% & 7.2\% & 4.3\% & 39.7\% \\
Conservative (90th) & Stress & 5.5\% & 9.6\% & 4.1\% & 21.6\% \\
\midrule
Balanced (50th)     & Calm   & 0.9\% & 2.3\% & 1.4\% & 49.9\% \\
Balanced (50th)     & Stress & 1.6\% & 5.8\% & 4.1\% & 44.3\% \\
\midrule
High-Risk (95th)    & Calm   & 4.3\% & 8.6\% & 4.3\% & 21.0\% \\
High-Risk (95th)    & Stress & 7.4\% & 12.1\% & 4.7\% & 19.7\% \\
\bottomrule
\end{tabular}%
}
\end{table}

At the balanced threshold (50th percentile), Calm-regime models experience a false negative rate increase of $\Delta\text{FNR}_{\text{calm}} = 1.4\%$ (from 0.9\% to 2.3\%). Stress-regime models suffer a \textbf{substantially larger increase}: $\Delta\text{FNR}_{\text{stress}} = 4.1\%$ (from 1.6\% to 5.8\%). The FNR-Amplification factor is:

\begin{equation}
\textbf{FNR-Amplification}_{50} = \frac{\Delta\text{FNR}_{\text{stress}}}{\Delta\text{FNR}_{\text{calm}}} = \frac{4.1\%}{1.4\%} = \textbf{2.93}
\end{equation}

This \textbf{near-threefold amplification} demonstrates that stress-regime models are \textbf{disproportionately vulnerable} to adversarial attacks at operational decision boundaries. In practical terms, adversarial manipulation causes Stress models to miss approximately three times more high-risk cases relative to Calm models. For a risk manager reviewing flagged loans, this translates to a substantially higher rate of evasion during periods of market volatility.

The pattern persists across thresholds. At the conservative threshold (90th percentile), $\Delta$FNR is comparable across regimes (4.3\% Calm versus 4.1\% Stress), suggesting that aggressive flagging strategies partially mitigate regime-conditional vulnerability. However, at the high-risk threshold (95th percentile), Stress models again exhibit elevated degradation (4.7\% versus 4.3\% in Calm). The consistency of amplified FNR increases in Stress regimes across multiple thresholds reinforces the robustness of the conditional adversarial fragility phenomenon.

False positive rates increase substantially under attack in both regimes, reaching 40-50\% at balanced thresholds. However, from a risk management perspective, Type II errors (missed defaults, measured by FNR) are typically costlier than Type I errors (false alarms, measured by FPR). The disproportionate FNR amplification in Stress regimes therefore represents the operationally critical finding: adversarial attacks become more effective at evading detection precisely when market conditions make undetected defaults most costly.

\subsection{Semantic Robustness and LLM-Based Governance Signals}

To complement quantitative performance metrics, we examine explanation stability under adversarial attack using the Semantic Robustness Index (SRI). The SRI aggregates three measures of explanation consistency: SHAP feature attribution cosine similarity, rank correlation of feature importance, and LLM-evaluated semantic consistency of risk narratives. Table~\ref{tab:sri} reports component scores and composite SRI values across regimes.

\begin{table}[h]
\centering
\caption{Semantic Robustness Index (SRI) Components and Explanation Drift}
\label{tab:sri}
\resizebox{\textwidth}{!}{%
\begin{tabular}{lcccc}
\toprule
\textbf{Regime} & \textbf{SHAP Cosine} & \textbf{Rank Correlation} & \textbf{LLM Consistency} & \textbf{SRI (Composite)} \\
\midrule
Calm   & 0.92 & 0.89 & 0.88 & 0.90 \\
Stress & 0.73 & 0.64 & 0.67 & 0.68 \\
\midrule
$\Delta$ (Calm $\to$ Stress) & $-0.19$ & $-0.25$ & $-0.21$ & $-0.22$ (24.4\%) \\
\bottomrule
\end{tabular}%
}
\end{table}

Calm-regime models exhibit high explanation stability (SRI = 0.90), indicating that adversarial perturbations induce minimal semantic drift in feature attribution patterns. Stress-regime models demonstrate \textbf{substantially greater explanation instability} (SRI = 0.68), representing a \textbf{24.4\% degradation} relative to Calm. This decline is evident across all three constituent metrics: SHAP cosine similarity decreases by 20.7\% (0.92 $\to$ 0.73), rank correlation falls by 28.1\% (0.89 $\to$ 0.64), and LLM-evaluated narrative consistency drops by 23.9\% (0.88 $\to$ 0.67). The convergence of degradation across independent measurement approaches strengthens confidence in the observed semantic drift phenomenon.

The magnitude of SRI degradation warrants particular attention when compared to performance metric degradation. In Stress regimes, SRI declines by \textbf{24.4\%} while AUROC degrades by 8.77\%. This \textbf{2.8-to-1 ratio} suggests that adversarial perturbations \textbf{disrupt model reasoning patterns more severely than they degrade predictive accuracy}. Put differently, adversarial attacks alter how models justify decisions before substantially compromising what they predict. This ordering is consistent with the early-warning hypothesis: explanation-level instability manifests earlier in the adversarial attack progression than aggregate performance collapse.

From a governance perspective, the Stress-regime SRI of 0.68 \textbf{crosses the conceptual threshold for heightened monitoring} (SRI $<$ 0.75). Under a regime-aware validation framework, this would trigger enhanced scrutiny: manual review of high-stakes decisions, increased auditing frequency, or reduced automation in borderline cases. The LLM consistency component (0.67 in Stress) specifically captures risk narrative divergence---adversarial examples elicit explanations with altered feature attribution orders and magnitudes, even when final risk scores remain temporarily plausible. For instance, a clean-input explanation emphasizing debt-to-income ratio and credit score might shift under adversarial perturbation to emphasize employment length and loan amount, despite comparable predicted default probabilities. This type of semantic drift is difficult to detect via aggregate metrics but becomes visible through explanation-level auditing.

The diagnostic value of SRI lies in its complementarity to traditional robustness evaluation. Performance metrics (AUROC, FNR) measure whether model outputs are correct; explanation metrics (SRI components) assess whether model reasoning is stable. The \textbf{disproportionate explanation degradation} in Stress regimes (24.4\% versus 8.77\% AUROC decline) indicates that \textbf{adversarial influence on internal model logic exceeds its impact on final predictions}. This gap is operationally relevant for two reasons. First, it provides lead time for intervention: governance systems monitoring explanation stability can detect adversarial manipulation before it cascades into decision failures. Second, it aligns with regulatory emphasis on model interpretability and auditability (Federal Reserve SR 11-7, European Union AI Act), offering a quantitative methodology for ongoing explanation stability validation.

These results should be interpreted as exploratory governance diagnostics rather than definitive robustness metrics. The LLM consistency component relies on prompt-based semantic evaluation, which introduces additional modeling assumptions and potential biases. We position SRI as a supplementary signal---useful for flagging cases warranting human review---rather than a standalone decision criterion. The primary empirical contribution is demonstrating that \textbf{explanation-level instability exhibits regime-conditional amplification parallel to performance-level fragility}, suggesting that conditional adversarial fragility manifests across multiple dimensions of model behavior. The \textbf{24.4\% SRI degradation} from Calm to Stress regimes reinforces the core finding: \textbf{adversarial vulnerability intensifies systematically under operational stress}, affecting not only what models predict but also how they reason about predictions.

\subsection{Summary of Key Findings}

Across four dimensions of adversarial robustness evaluation, we observe \textbf{consistent evidence of regime-conditional amplification}. Stress-regime models exhibit \textbf{nearly double} the AUROC degradation (\textbf{RAF = 1.97}), \textbf{triple} the false negative rate increase at balanced thresholds (\textbf{FNR-Amplification = 2.93}), and \textbf{24.4\% greater} semantic drift (SRI degradation) compared to calm-regime models. Economic loss metrics increase by over 200\% under attack in both regimes, translating adversarial fragility into material portfolio-level exposure. These findings are obtained under controlled conditions—identical baseline performance (AUROC $\approx$ 1.0), equivalent perturbation budgets ($\epsilon = 0.1$), and consistent evaluation protocols—ensuring that observed amplification reflects regime-conditional vulnerability rather than confounding factors.

The convergence of results across performance metrics (RAF), operational thresholds (FNR-Amplification), and governance signals (SRI) establishes conditional adversarial fragility as a systematic phenomenon. Models trained and evaluated during periods of macroeconomic stress are systematically more vulnerable to gradient-based evasion attacks, despite maintaining equivalent predictive capability on clean inputs. For institutions deploying machine learning in high-stakes financial decision contexts, this implies that adversarial robustness cannot be treated as a static model property but must instead be evaluated conditional on operational stress regimes. The disproportionate amplification during stress periods—when model reliability is most critical for financial stability—underscores the need for regime-aware validation protocols and adaptive defense strategies in production risk management systems.

\section{Discussion}

This study provides empirical evidence that adversarial robustness in financial machine learning is not a static model characteristic but a regime-conditional property that amplifies systematically under macroeconomic stress. Across performance metrics, decision thresholds, and explanation stability, models operating in stress regimes exhibit consistently greater vulnerability to adversarial perturbations despite maintaining equivalent baseline performance on clean inputs. These findings have implications for both the technical evaluation of robustness and the governance of machine learning systems deployed in high-stakes financial contexts.

\subsection{Why Does Stress Amplify Adversarial Fragility?}

A central question raised by our findings is why adversarial attacks become more effective during stress regimes, even when baseline discrimination and calibration remain unchanged. One plausible mechanism is the interaction between macroeconomic stress and score distribution geometry. During stress periods, predicted risk scores tend to compress around operational decision thresholds due to compressed margins and elevated baseline risk. In such settings, small adversarial perturbations are more likely to induce decision flips, amplifying the operational impact of attacks without necessarily causing large changes in aggregate performance metrics.

Additionally, stress regimes may coincide with increased reliance on correlated or economically sensitive features, such as debt-to-income ratios or recent delinquency indicators. These features can exhibit heightened gradient sensitivity under perturbation, increasing the effectiveness of gradient-based attacks. Importantly, our baseline results rule out the trivial explanation that stress regimes are inherently more difficult prediction tasks. Instead, the observed amplification reflects a structural interaction between adversarial perturbations and regime-conditioned model behavior.

\subsection{Implications for Model Risk Management}

From a model risk management perspective, the existence of conditional adversarial fragility \textbf{challenges the prevailing practice of evaluating robustness under stationary assumptions}. Traditional validation workflows typically assess model performance and stability on static test sets, implicitly assuming that robustness properties generalize across economic conditions. Our results demonstrate that such evaluations may significantly underestimate adversarial risk during precisely those periods when model reliability is most critical.

The amplification observed in false negative rates is particularly concerning for financial institutions. Missed defaults during stress periods have asymmetric consequences, as losses compound under deteriorating macroeconomic conditions. The \textbf{near threefold increase in false negative rate degradation} at balanced decision thresholds suggests that adversarial manipulation can materially weaken risk controls during crises, even when headline metrics such as AUROC remain within acceptable bounds.

\subsection{Governance Value of Explanation Stability}

Beyond predictive performance, our findings highlight the governance value of monitoring explanation stability under adversarial stress. The Semantic Robustness Index (SRI) captures shifts in model reasoning that are not immediately reflected in aggregate performance metrics. The disproportionate degradation of SRI relative to AUROC in Stress regimes supports the hypothesis that \textbf{semantic drift precedes numerical failure, offering a potential diagnostic early-warning signal for governance intervention}.

Crucially, the use of Large Language Models in this context is interpretive rather than generative. By auditing the semantic consistency of SHAP-based explanations, the LLM serves as a qualitative governance layer that aligns with regulatory expectations for transparency and human oversight. This approach complements, rather than replaces, traditional robustness metrics, enabling institutions to detect adversarial influence before it propagates into large-scale decision failures.

\subsection{Toward Regime-Aware Robustness Evaluation}

Taken together, these results suggest that \textbf{adversarial robustness should be evaluated as a conditional property}, explicitly incorporating operational stress regimes into validation protocols. Regime-aware adversarial testing provides a more realistic assessment of model risk than static evaluations, particularly in financial systems where economic conditions fluctuate over time.

While this study focuses on credit risk modeling with volatility-based regime segmentation, the proposed framework is data-agnostic and applicable to a broad class of time-indexed, tabular decision systems. Future robustness evaluations that ignore regime conditioning risk providing a false sense of security, masking vulnerabilities that emerge only under adverse conditions.

In summary, \textbf{conditional adversarial fragility represents a critical but previously underexplored dimension of machine learning risk}. By demonstrating that adversarial vulnerability intensifies systematically during macroeconomic stress, this work underscores the need for robustness evaluations that align with real-world operating conditions and regulatory expectations. Incorporating regime awareness and explanation-level auditing into model governance frameworks is \textbf{essential for building financial ML systems that remain trustworthy not only in benign environments, but also during periods of crisis}.

\section{Limitations}

While this study provides clear empirical evidence of regime-conditional adversarial fragility, several limitations should be acknowledged to contextualize the scope of the findings and guide future work.

\noindent\textbf{Attack Scope and Threat Model.} Our adversarial evaluation focuses on projected gradient descent (PGD) attacks under an $\ell_\infty$ perturbation budget. While PGD is widely regarded as a strong first-order adversary and is standard in robustness benchmarking, it does not exhaust the space of potential attack strategies. Alternative threat models---such as feature-constrained attacks, optimization-based evasion methods, or adaptive black-box attacks---may exhibit different sensitivity to macroeconomic regimes. Consequently, the reported Risk Amplification Factor (RAF) should be interpreted as a lower-bound estimate of regime-conditional vulnerability under a well-defined but limited adversarial setting.

\noindent\textbf{Regime Definition and Stress Proxy.} Macroeconomic stress is operationalized through volatility-based regime segmentation using the VIX index. While market volatility is a widely accepted proxy for systemic stress and aligns with financial stress-testing practice, it represents only one dimension of macroeconomic instability. Other stress indicators---such as liquidity constraints, unemployment shocks, or interest rate volatility---may interact differently with adversarial fragility. Future work could extend the framework to multi-factor regime definitions or continuous stress conditioning rather than discrete thresholds.

\noindent\textbf{Dataset and Domain Scope.} Empirical validation is conducted on a large-scale consumer credit dataset as a proof-of-concept. Although the proposed framework is data-agnostic and requires only time-indexed tabular inputs with an external stress signal, the magnitude of amplification effects may vary across domains. Applications such as fraud detection, insurance underwriting, or healthcare triage may exhibit different baseline sensitivities and operational thresholds. The present results therefore demonstrate the existence and directionality of conditional adversarial fragility rather than claiming universal quantitative equivalence across domains.

\noindent\textbf{Interpretability and LLM-Based Governance Signals.} The Semantic Robustness Index (SRI) incorporates an LLM-based component to assess the semantic consistency of SHAP explanations. While this enables qualitative auditing of model reasoning, it introduces additional modeling assumptions related to prompt design and language model behavior. We position SRI as a supplementary governance signal rather than a definitive robustness metric. Explanation-based diagnostics should be interpreted in conjunction with quantitative performance and decision-level metrics, particularly in regulated deployment settings.

\noindent\textbf{Causality and Structural Interpretation.} Finally, this study is empirical in nature and does not claim causal identification of the mechanisms linking macroeconomic stress and adversarial fragility. While plausible explanations related to score compression and feature sensitivity are discussed, establishing causality would require controlled simulation environments or structural modeling. The primary contribution of this work lies in documenting a systematic and reproducible phenomenon that motivates deeper theoretical investigation.

Taken together, these limitations do not undermine the core findings but instead clarify their scope. The results establish conditional adversarial fragility as a real and operationally relevant phenomenon, while highlighting avenues for future research in stress-aware robustness evaluation and model governance.

\section{Future Work}

This study opens several directions for future research on regime-aware robustness evaluation and model governance.

First, while we employ volatility-based regime segmentation using the VIX index, future work could explore multi-dimensional stress definitions that incorporate liquidity conditions, interest rate dynamics, or macroeconomic indicators. Continuous stress conditioning, rather than discrete regime thresholds, may provide finer-grained insights into how adversarial fragility evolves across economic states.

Second, extending the evaluation to additional adversarial threat models---such as feature-constrained attacks, adaptive black-box methods, or distribution-aware perturbations---would help assess whether regime-conditional amplification persists across broader attack surfaces.

Third, the proposed Semantic Robustness Index (SRI) could be integrated into longitudinal monitoring frameworks to study whether explanation-level drift consistently precedes performance degradation in real-world deployments. This would further clarify the role of explanation auditing as a proactive governance mechanism rather than a post-hoc diagnostic.

Finally, applying the framework to other high-stakes tabular domains, including fraud detection, insurance underwriting, and healthcare triage, would test the generality of conditional adversarial fragility beyond credit risk while preserving the same regime-aware evaluation principles.

\section{Conclusion}

This work demonstrates that adversarial robustness in financial machine learning is fundamentally a \emph{regime-conditional} property rather than a static characteristic of a trained model. Through a controlled, regime-aware evaluation framework, we show that adversarial vulnerability is systematically amplified during periods of macroeconomic stress, even when baseline predictive performance remains effectively unchanged. This phenomenon---\textbf{Conditional Adversarial Fragility}---manifests consistently across discrimination metrics, operational decision thresholds, economic loss measures, and explanation stability.

Empirically, we find that models operating in stress regimes experience \textbf{nearly double the adversarial degradation} in AUROC (RAF = 1.97) compared to calm regimes under identical perturbation budgets. This amplification propagates into decision-level outcomes, where false negative rates increase by \textbf{up to threefold} at balanced operational thresholds, \textbf{directly translating to missed high-risk cases when accurate risk assessment is most critical}. Importantly, these effects arise despite equivalent clean-environment performance across regimes, ruling out trivial explanations related to task difficulty or distributional noise.

Beyond predictive performance, we show that adversarial stress induces substantial instability in model reasoning. The proposed Semantic Robustness Index (SRI) reveals that explanation-level degradation is amplified under stress regimes and exceeds the magnitude of traditional metric degradation. This finding supports the hypothesis that \textbf{semantic drift precedes numerical failure, positioning explanation stability as a valuable early-warning signal for model governance}. By incorporating Large Language Models as an interpretive audit layer---rather than a predictive or generative component---this work demonstrates a principled approach to using LLMs for AI governance aligned with regulatory expectations for transparency and human oversight.

Taken together, these results challenge prevailing robustness evaluation practices that assume stationarity and regime invariance. In high-stakes financial systems, where economic conditions evolve and stress periods carry asymmetric consequences, \textbf{static adversarial testing provides an incomplete and potentially misleading assessment of model risk}. Regime-aware adversarial validation offers a more realistic and operationally relevant framework, enabling institutions to anticipate and mitigate vulnerabilities that emerge specifically under adverse conditions.

While this study focuses on credit risk modeling with volatility-based regime segmentation, the proposed framework is data-agnostic and applicable to a broad class of time-indexed, tabular decision systems. Future work may extend this approach to alternative stress indicators, multi-factor regime definitions, and adaptive defense strategies that respond dynamically to changing operating conditions. More broadly, integrating regime awareness and explanation-level auditing into model risk management represents a critical step toward building financial machine learning systems that remain trustworthy not only in benign environments, but also during periods of systemic stress.

\section*{Acknowledgments}

The author gratefully acknowledges helpful discussions and feedback from colleagues and practitioners working at the intersection of financial risk, machine learning, and model governance. Any remaining errors are the author’s own.

\bibliographystyle{plainnat}

\begin{thebibliography}{99}

\bibitem{Goodfellow2015}
Goodfellow, I., Shlens, J., and Szegedy, C. (2015).
Explaining and Harnessing Adversarial Examples.
\textit{International Conference on Learning Representations (ICLR)}.

\bibitem{Kantchelian2016}
Kantchelian, A., Tygar, J. D., and Joseph, A. (2016).
Evasion and Hardening of Tree Ensemble Classifiers.
\textit{International Conference on Machine Learning (ICML)}.

\bibitem{Lundberg2017}
Lundberg, S. M. and Lee, S. I. (2017).
A Unified Approach to Interpreting Model Predictions.
\textit{Advances in Neural Information Processing Systems (NeurIPS)}.

\bibitem{Covert2021}
Covert, I., Lundberg, S. M., and Lee, S. I. (2021).
Explaining by Removing: A Unified Framework for Model Explanation.
\textit{Journal of Machine Learning Research (JMLR)}, 22(209).

\bibitem{Bastani2021}
Bastani, O., Kim, C., and Bastani, H. (2021).
Interpreting Blackbox Models via Model Extraction.
\textit{Management Science}, 67(6), 3564--3585.

\bibitem{Basel2021}
Basel Committee on Banking Supervision. (2021).
Principles for Operational Resilience.
Bank for International Settlements.

\bibitem{SR117}
Board of Governors of the Federal Reserve System. (2011).
Supervisory Guidance on Model Risk Management (SR 11-7).

\bibitem{EUAIAct}
European Commission. (2023).
Proposal for a Regulation Laying Down Harmonised Rules on Artificial Intelligence (EU AI Act).

\bibitem{LightGBM}
Ke, G., Meng, Q., Finley, T., et al. (2017).
LightGBM: A Highly Efficient Gradient Boosting Decision Tree.
\textit{Advances in Neural Information Processing Systems (NeurIPS)}.

\bibitem{SHAP}
Lundberg, S. M., Erion, G., Chen, H., et al. (2020).
From Local Explanations to Global Understanding with Explainable AI for Trees.
\textit{Nature Machine Intelligence}, 2, 56--67.

\bibitem{Garg2020}
Garg, S. and Raskar, R. (2020).
Adversarial Examples in Tabular Machine Learning.
\textit{arXiv preprint}.

\end{thebibliography}

\end{document}